\documentclass[10pt,twocolumn,letterpaper]{article}

\usepackage{cvpr}
\usepackage{times}
\usepackage{epsfig}
\usepackage{graphicx}
\usepackage{amsmath}
\usepackage{amssymb}
\usepackage{xcolor}
\usepackage{multirow,tabularx}
\usepackage[percent]{overpic}

\usepackage[pagebackref=true,breaklinks=true,letterpaper=true,colorlinks,bookmarks=false]{hyperref}

\newcommand\blfootnote[1]{%
	\begingroup
	\renewcommand\thefootnote{}\footnote{#1}%
	\addtocounter{footnote}{-1}%
	\endgroup
}

\renewcommand{\vec}[1]{\boldsymbol{#1}}
\newcommand{\mat}[1]{\mathbf{#1}}
\newcommand{\set}[1]{\mathcal{#1}}

\newcommand{\smpl}[0]{M}
\newcommand{\posefun}[0]{T}
\newcommand{\blendfun}[0]{W}
\newcommand{\offsetfun}[0]{B}
\newcommand{\jointfun}[0]{J}

\newcommand{\pose}[0]{\vec{\theta}}
\newcommand{\shape}[0]{\vec{\beta}}
\newcommand{\trans}[0]{\vec{t}}
\newcommand{\joints}[0]{\mat{J}}

\newcommand{\blendweights}[0]{\mat{W}}
\newcommand{\template}[0]{\mat{T}}
\newcommand{\offsets}[0]{\mat{D}}

\newcommand{\Jopenpose}[0]{J_\text{B25}}

\newcommand{\imageset}[0]{\set{I}}
\newcommand{\image}[0]{\mat{I}}

\newcommand{\poseset}[0]{\set{P}}
\newcommand{\transset}[0]{\set{T}}

\newcommand{\net}[0]{f_w}
\newcommand{\netstar}[0]{f_w^{*}}

\newcommand{\outshapetpose}[0]{S}
\newcommand{\outshapeposed}[0]{N}

\newcommand{\loss}[0]{\mathcal{L}}

\newcommand{\latentCode}[0]{l^{\text{inv}}}
\newcommand{\latentPose}[0]{l^{\text{pose}}}

\cvprfinalcopy %

\ifcvprfinal\fi

\begin{document}

\title{Learning to Reconstruct People in Clothing from a Single RGB Camera}

\author{Thiemo Alldieck\textsuperscript{1,2,*}\hspace{8mm}
	Marcus Magnor\textsuperscript{1}\hspace{8mm}
	Bharat Lal Bhatnagar\textsuperscript{2}\\
	Christian Theobalt\textsuperscript{2}\hspace{8mm}
	Gerard Pons-Moll\textsuperscript{2}
}

\makeatletter
\let\@oldmaketitle\@maketitle%
\renewcommand{\@maketitle}{
    \@oldmaketitle%
    \centering
    \vspace{-6mm}
   	{\small \textsuperscript{1}Computer Graphics Lab, TU Braunschweig, Germany}\\
    {\small	\textsuperscript{2}Max Planck Institute for Informatics, Saarland Informatics Campus, Germany}\\
    {\tt\scriptsize \{alldieck,magnor\}@cg.cs.tu-bs.de \{bbhatnag,theobalt,gpons\}@mpi-inf.mpg.de}\\
    \vspace{2mm}
    \centering
    \includegraphics[width=0.97\textwidth]{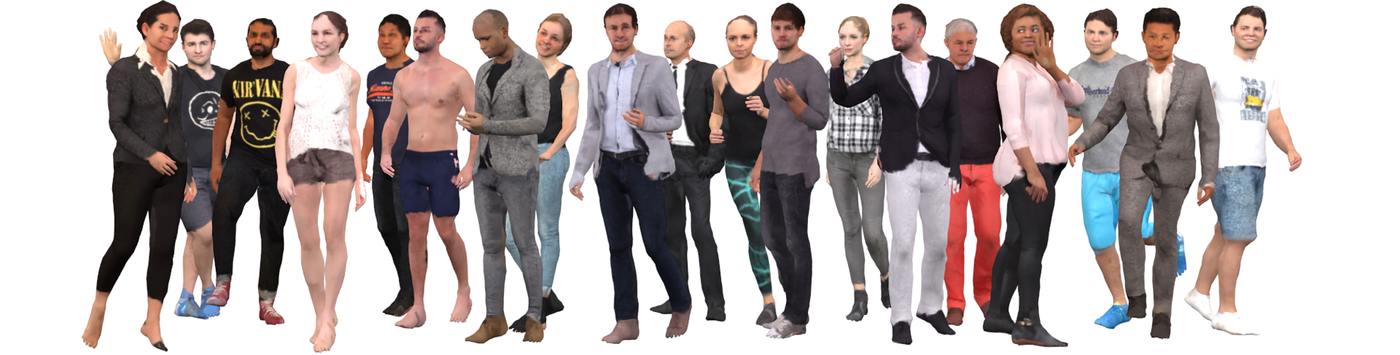}\\
    \refstepcounter{figure}\small Figure~\thefigure: We present a deep learning based approach to estimate personalized body shape, including hair and clothing, using a single RGB camera. The shapes shown above have been calculated using only 8 input images, and re-posed using SMPL.
    \label{fig:teaser}
    \vspace{4mm}
}
\makeatother

\maketitle
\thispagestyle{plain}
\pagestyle{plain}

\begin{abstract}
We present Octopus, a learning-based model to infer the personalized 3D shape of people from a few frames (1-8) of a monocular video in which the person is moving with a reconstruction accuracy of 4 to 5mm, while being orders of magnitude faster than previous methods. From semantic segmentation images, our Octopus model reconstructs a 3D shape, including the parameters of SMPL plus clothing and hair in 10 seconds or less.  The model achieves fast and accurate predictions based on two key design choices. First, by predicting shape  in a canonical T-pose space, the network learns to encode the images of the person into pose-invariant latent codes, where the information is fused. Second, based on the observation that feed-forward predictions are fast but do not always align with the input images, we predict using both, bottom-up and top-down streams (one per view) allowing information to flow in both directions. Learning relies only on synthetic 3D data.
Once learned, Octopus can take a variable number of frames as input, and is able to reconstruct shapes even from a single image with an accuracy of 5mm. Results on 3 different datasets demonstrate the efficacy and accuracy of our approach. Code is available at \cite{code}.
\end{abstract}
\blfootnote{* Work partly conducted during an internship at the Real Virtual Humans group of Max Planck for Informatics.}

\section{Introduction}
\label{sec:introduction}
The automatic acquisition of detailed 3D human shape and appearance, including clothing and facial details is required for many applications such as VR/AR, gaming, virtual try-on, and cinematography.

A common way to acquire such models is with a scanner or a multi-view studio~\cite{Ahmed:2005:VRST2005,LWSLVDT13}. The cost and size prevent the wide-spread use of such setups. Therefore, numerous works address capturing body shape and pose with more practical setups, e.g. from a low number of video cameras~\cite{rhodin2016general}, or using one or more depth cameras, either specifically for the human body~\cite{Bogo:ICCV:2015,weiss2011home,zhang2014quality} or for general free-form surfaces~\cite{zollhofer2014real,newcombe2015dynamicfusion,orts2016holoportation,innmann2016volume,dou2016fusion4d,DoubleFusion2018}. The most practical but also challenging setting is capturing from a single monocular RGB camera. Some methods attempt to infer the shape parameters of a body model from a single image~\cite{kanazawa2018endtoend,omran2018neural,bogo2016smplify,dibra2017human,bualan2008naked,hasler2010multilinear,zhou2010parametric,jain2010moviereshape,pavlakos2018humanshape}, 
but reconstructed detail is constrained to the model shape space, and thus does not capture personalized shape detail and clothing geometry.
Recent work~\cite{alldieck2018video,alldieck2018detailed} estimates more detailed shape, including clothing, from a video sequence of a person rotating in front of a camera while holding a rough A-pose. %
While reconstructed models have high quality, the optimization approach takes around $2$ minutes only for the shape component.  More importantly, the main bottleneck is the pre-processing step, which requires \emph{fitting} the SMPL model to each of the frame silhouettes using time-consuming non-linear optimization ($\approx120$~min for $120$ frames). This is impractical for many applications that require fast acquisition such as telepresence and gaming.

In this work, we address these limitations and introduce \emph{Octopus}, a convolutional neural network (CNN) based model that learns to predict 3D human shapes in a canonical pose given a few frames of a person rotating in front of a \emph{single camera}. Octopus predicts using both, bottom-up and top-down streams (one per view) allowing information to flow in both directions. It can make bottom-up predictions in $50$ms per view, which are effectively refined top-down using the same images in $10$s. Inference, both bottom-up and top-down, is performed fully-automatically using the same model. \emph{Octopus} is therefore easy to use and more practical than previous work~\cite{alldieck2018video}. Learning only relies on synthetic 3D data, and on semantic segmentation images and keypoints derived from synthesized video sequences. Consequently, \emph{Octopus} can be trained without paired data -- real images with ground truth 3D shape annotations -- which is very difficult to obtain in practice. 

Octopus predicts SMPL body model parameters, which represent the undressed shape and the pose, plus additional 3D vertex offsets that model clothing, hair, and details beyond the SMPL space. Specifically, a CNN encodes $F$ frames of the person (in different poses) into $F$ latent codes that are fused to obtain a single shape code. From the shape code, two separate network streams predict the SMPL shape parameters, and the 3D vertex offsets in the canonical T-pose space, giving us the ``unpose'' shape or T-shape.  
Predicting the T-shape forces the $F$ latent codes to be pose-invariant, which is necessary to fuse the shape information contained in each frame. Octopus also predicts a pose for each frame, which allows to ``pose'' the T-shape and render a silhouette to evaluate the overlap against the input images in a top-down manner during both training and inference. Specifically, since bottom-up models do not have a feedback loop, the feed-forward 3D predictions are correct but do not perfectly align with the input images. Consequently, we refine the prediction top-down by optimizing the $F$ poses, the T-shape, and the vertex offsets to
maximize silhouette overlap and joint re-projection error.

Experiments on a newly collected dataset (LifeScans), the publicly available PeopleSnapshot dataset~\cite{alldieck2018video}, and on the dataset used in~\cite{Bogo:ICCV:2015} demonstrate that our model infers shapes with a reconstruction accuracy of $4$mm in less than $10$ seconds.
 In summary, 
 \emph{Octopus} is faster than purely optimization-based fitting approaches such as~\cite{alldieck2018video}, it combines the advantages of bottom-up and top-down methods in a single model, and can reconstruct detailed shapes and clothing from a few video frames. Examples of reconstruction results are shown in Fig.~\ref{fig:teaser}. To foster further research in this direction, we made Octopus available for research purposes~\cite{code}.

\section{Related Work}

\label{sec:related}
\vspace{-1mm}

Methods for 3D human shape and pose reconstruction can be broadly classified as top-down or bottom-up.
Top-down methods either fit a \emph{free-form} surface or a  statistical body model (\emph{model-based}).
Bottom-up methods directly infer a surface or body model parametrization from sensor data.
We will review bottom-up and top-down methods for human reconstruction.

\vspace{-4.5mm}
\paragraph{Top-down, free-form} methods non-rigidly deform meshes~\cite{carranza2003free,deAguiar2008performance,cagniart_meshdeform} or volumetric shape representations~\cite{huang2016volumetric,InriaVolumetric_2015}. These methods are based on multi-view stereo reconstruction~\cite{koch1998multi}, and therefore require multiple RGB or depth cameras, which is a practical barrier for many applications.
Using \emph{depth cameras}, KinectFusion~\cite{izadi2011kinectfusion,newcombe2011kinectfusion} approaches reconstruct 3D scenes by incrementally fusing frame geometry, and appearance~\cite{zhou2014color}, in a canonical frame. Several methods build on KinectFusion for body scanning~\cite{shapiro2014rapid,3Dportraits,zeng2013templateless,cui2012kinectavatar}. The problem is that these methods require the person to stand still while the camera is turned around.  
DynamicFusion~\cite{newcombe2015dynamicfusion} generalized KinectFusion to non-rigid objects by combining non-rigid tracking and fusion. Although template-free approaches~\cite{newcombe2011kinectfusion,innmann2016volume,slavcheva2017killingfusion} are flexible, they can only handle very careful motions. 
Common ways to add robustness are pre-scanning the template~\cite{zollhofer2014real}, or using multiple kinects~\cite{dou2016fusion4d,orts2016holoportation} or multi-view~\cite{starck2007surface,inria_2017,collet2015high}.
These methods, however, do not register the temporal 3D reconstructions to the same template and focus on other applications such as streaming or telepresence~\cite{orts2016holoportation}.
Estimating shape by compensating for pose changes can be tracked back to Cheung~\emph{et al.}~\cite{cheung2003shape,cheung2003visual}, where they align visual hulls over time to improve shape estimation. To compensate for articulation, they merge shape information in a coarse voxel model. However, they need to track each body part separately and require multi-view input. All free-form works require multi-view input, depth cameras or cannot handle moving humans.

\vspace{-4.5mm}
\paragraph{Top-down, model-based} methods exploit a parametric body model consisting of pose and shape~\cite{anguelov2005scape,hasler2009statistical,smpl2015loper,zuffi2015stitched,pons2015dyna,joo2018total} to regularize the fitting process. 
Some ~\emph{Depth-based} methods~\cite{weiss2011home,Helten:2013,ye2014real,zhang2014quality,Bogo:ICCV:2015} exploit the temporal information by optimizing a single shape and multiple poses (jointly or sequentially). This leads to expensive optimization problems. Using~\emph{mutli-view}, some works achieve fast performance~\cite{rhodin2016general,Robertini:2016} at the cost of using a coarser body model based on Gaussians~\cite{stoll2011fast}, or a pre-computed template~\cite{Yu_2015_ICCV}. Early RGB-based methods were restricted to estimating the parameters of a body model, and required multiple views~\cite{bualan2008naked} or manually clicked points~\cite{guan2009estimating,zhou2010parametric,jain2010moviereshape,rogge2014garment}. Shape and clothing have been recovered from RGB images~\cite{guo2012clothed,chen2013deformable},  depth~\cite{chen2015garment}, or scan data \cite{ponsmoll2017clothcap}, but require manual intervention or clothing is limited to a pre-defined set of templates. 
In~\cite{yang2018analyzing} a fuzzy vertex association from clothing to body surface is introduced, which allows complex clothing modeled as body offsets.
Some works are in-between free-form and model-based methods. In~\cite{gall2009motion,vlasic2008articulated}, authors pre-scan a template and insert a skeleton, and in~\cite{DoubleFusion2018} authors combine the SMPL model with a volumetric representation to track the clothed human body from a depth camera. 

\vspace{-4.5mm}
\paragraph{Bottom-up.}
Learning of features for multi-view photo-consistency~\cite{leroy2018shape}, and auto-encoders combined with visual hulls~\cite{gilbert2018volumetric, Trumble2018deepAutoEncoder} have shown to improve free-form performance capture. These works, however, require more than one camera view. 
Very few works learn to predict personalized human shape from images--lack of training data and the lack of a feedback loop between feed-forward predictions and the images makes the problem hard.
Variants of random forests and neural networks have been used~\cite{dibra2017human,dibra2016hs,dibra2016shape, varol17_surreal} to regress shape from silhouettes. The problem here is that predictions tend to look over-smooth, are confined to the model shape space, and do not comprise clothing. Garments are predicted~\cite{danvevrek2017deepgarment} from a single image, but a single model for every new garment needs to be trained, which makes it hard to use in practice.
Recent pure bottom-up approaches to human analysis~\cite{mehta2017vnect,mehta2018multiperson,popa2017deep, zhou2017towards,sun2017compositional,tome2017lifting,rogez_lcr_cvpr17} typically predict shape represented as a coarse stick figure or bone skeleton, and can not estimate body shape or clothing.

\vspace{-4.5mm}
\paragraph{Hybrid methods.}
A recent trend of works combines bottom-up and top-down approaches--a combination that has been exploited already in earlier works~\cite{sminchisescu2006learning}. %
The most straightforward way is by fitting a 3D body model~\cite{smpl2015loper} to 2D pose detections~\cite{bogo2016smplify,Lassner}.
These methods, however, can not capture clothing and details beyond the model space. Clothing, hair and shape~\cite{alldieck2018video,alldieck2018detailed} can be inferred by fusing dynamic silhouettes (predicted bottom-up) of a video to a canonical space. Even with good 2D predictions,
these methods are susceptible to local minima when not initialized properly, and are typically slow. Furthermore, the 2D prediction network and the model fitting is de-coupled. 
Starting with a feed-forward 3D prediction, semantic segmentation, keypoints and scene constraints are integrated top-down in order to predict the pose and shape of multiple people~\cite{zanfir2018monocular}. Other recent works integrate the SMPL model, or a voxel representation~\cite{varol2018bodynet}, as a layer within a network architecture~\cite{kanazawa2018endtoend,pavlakos2018humanshape,omran2018neural,tung2017self}. This has several advantages: (i) predictions are constrained by a shape space of humans, and (ii) bottom-up 3D predictions can be verified top-down using 2D keypoints and silhouettes during training. However, the shape estimates are confined to the model shape space and tend to be close to the average. The focus of these works is rather on robust pose estimation, while we focus on personalized shapes.   
We also integrate SMPL within our architecture but our work is different in several aspects.
First, our architecture fuses the information of several images of the same person in different poses. Second, our model incorporates a fast top-down component during training \emph{and} at test time. As a result, we can predict clothing, hair and personalized shapes using a single camera.

\section{Method}
\label{sec:method}

\begin{figure*}
	\centering
	\begin{overpic}[width=0.95\textwidth]{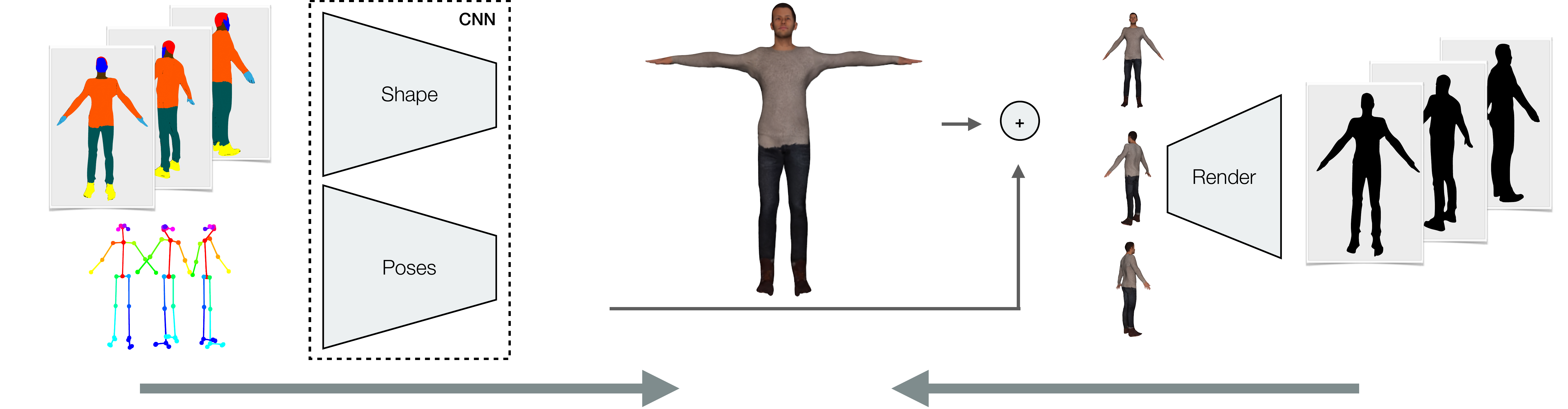}		%
		\put (34,19) {$\shape$, $\offsets$}
		\put (34,7.75) {$\poseset$, $\transset$}
		\put (66,20) {\small{$\pose_0$, $\trans_0$}}
		\put (66,12.5) {\small{$\pose_1$, $\trans_1$}}
		\put (66,5.5) {\small{$\pose_2$, $\trans_2$}}
		
		\put (1,17) {$\imageset$}
		\put (1,7) {$\set{J}$}
		\put (50,3) {\makebox(0,0){\footnotesize{$\outshapetpose(\imageset)$}}}
		\put (71,3) {\makebox(0,0){\footnotesize{$\outshapeposed_{3D}(\imageset,\set{J},i)$}}}
		\put (92,3) {\makebox(0,0){\footnotesize{$\outshapeposed_{2D}(\imageset,\set{J},i)$}}}
	\end{overpic}
	\caption{Overview of our method: Our novel CNN predicts 3D human shapes from semantic images in an canonical pose together with per-image pose information calculated from 2D joint detections (left to center). The pose information can be used to refine the shape via `render and compare' optimization using the same predictor (right to center).}
	\label{fig:overview}
	\vspace{-3mm}
\end{figure*}

The goal of this work is to create a 3D model of a subject from a few frames of a monocular RGB video, and in less than $10$ seconds.  
The model should comprise body shape, hair, and clothing and should be animatable. 
We take inspiration from~\cite{alldieck2018video} and focus on the cooperative setting with videos of people rotating in front of a camera holding a rough A-pose -- this motion is easy and fast to perform, and ensures that non-rigid motion of clothing and hair is not too large. 
In contrast to previous work~\cite{alldieck2018video}, we aim for fast and fully automatic reconstruction.
To this end, we train a novel convolutional neural network to infer a 3D mesh model of a subject from a small number of input frames. 
Additionally, we train the network to reconstruct the 3D pose of the subject in each frame.
This allows us to refine the body shape by utilizing the decoder part of the network for instance-specific optimization (Fig.~\ref{fig:overview}).

In Sec.~\ref{sec:smpl} we describe the shape representation used in this work followed by its integration into the used predictor (Sec.~\ref{sec:model_representation}).
In Sec.~\ref{sec:losses} we explain the losses, that are used in the experiments. We conclude by describing the instance-specific top-down refinement of results (Sec.~\ref{sec:Scene-specific optimization}).

\subsection{Shape representation}
\label{sec:smpl}
Similar to previous work~\cite{zahng2017shapeundercloth,alldieck2018video}, we represent shape using the SMPL statistical body model~\cite{smpl2015loper}, which represents the undressed body, and a set of offsets modeling instance specific details including clothing and hair.

SMPL is a function $\smpl(\cdot)$ that maps  pose $\pose$ and shape $\shape$ to a mesh of $V=6890$ vertices.
By adding offsets $\offsets$ to the template $\template$, we obtain a posed shape instance as follows: 
\vspace{-2mm}
\begin{equation}
\label{eq:smplpose}
\smpl(\shape,\pose,\offsets) = \blendfun(\posefun(\shape,\pose,\offsets), \jointfun(\shape), \pose, \blendweights)
\end{equation}
\begin{equation}
\label{eq:smplshape}
\posefun(\shape,\pose,\offsets) = \template + \offsetfun_s(\shape) + \offsetfun_p(\pose) + \offsets,
\end{equation}
where linear blend-skinning $\blendfun(\cdot)$ with weights $\blendweights$, together with pose-dependent deformations $\offsetfun_p(\pose)$ allow to pose the T-shape ($ \template + \offsetfun_s(\shape)$)  based on its skeleton joints $\jointfun(\cdot)$.
SMPL plus offsets, denoted as SMPL+D, is fully differentiable with respect to pose $\pose$, shape $\shape$ and free-form deformations $\offsets$.
This allows us to directly integrate SMPL as a fixed layer in our convolutional architecture.

\subsection{Model and data representation}
\label{sec:model_representation}
Given a set of images $\imageset=\{\image_0,\dots,\image_{F-1}\}$ depicting a subject from different sides with corresponding 2D joints $\set{J}=\{\joints_0,\dots,\joints_{F-1}\}$, we learn a predictor $\netstar$ that infers the body shape $\shape$, personal and scene specific body features $\offsets$, and 3D poses $\poseset=\{\pose_0,\dots,\pose_{F-1}\}$ along with 3D positions $\transset=\{\trans_0,\dots,\trans_{F-1}\}$ for each image.
$\netstar:(\imageset, \set{J} ) \mapsto (\shape, \offsets, \poseset, \transset)$ is a CNN parametrized by network parameters $w$.

\textbf{Input modalities.}
Images of humans are highly diverse in appearance, requiring large datasets of annotated images in the context of deep learning.
Therefore, to abstract away as much information as possible while still retaining shape and pose signal, we build on previous work~\cite{gong2018instance,cao2017realtime} to simplify each RGB image to a semantic segmentation and 2D keypoint detections. This allows us to train the network using only synthetic data and generalize to real data. 

\textbf{Model parametrization.}
By integrating the SMPL+D model (Sec.~\ref{sec:smpl}) into our network formulation, we can utilize its mesh output in the training of $\netstar$.
Concretely, we supervise predicted SMPL+D parameters in three ways:
Imposing a loss directly on the mesh vertices $M(\shape,\pose,\offsets)$, on the predicted joint locations $J(\shape)$ and their projections on the image, and densely on a rendering of the mesh using a differential renderer \cite{henderson18bmvc}.

The T-shape ($\template + \offsetfun_s(\shape) + \offsets$) in Eq.~\ref{eq:smplshape} is now predicted from the set of semantic images $\imageset$ with the function:
\begin{equation}
    \outshapetpose(\imageset) = \template + \offsetfun_s(\net^{\shape}(\imageset)) + \net^{\offsets}(\imageset),
\end{equation}
where $\netstar$ are the regressors to be learned.
Similarly, the mesh posed $\outshapeposed_{3D}(\imageset,\set{J}, i)$ is predicted from the image $\image_i$ and 2D joints $\joints_i$ with the function:
\begin{equation}
	\small
    \outshapeposed_{3D}(\imageset,\set{J}, i) = \blendfun(P(\imageset, i), \jointfun(\net^{\shape}(\imageset)),
    \net^{\pose_i}(\imageset,\set{J}), \blendweights)
\end{equation}
\begin{equation}
\small
    P(\imageset,\set{J}, i) = \outshapetpose(\imageset) + \offsetfun_p(\net^{\pose_i}(\imageset,\set{J})),
\end{equation}
from which the 3D Joints are predicted with the linear regressor $\Jopenpose$:
\begin{equation}
    \outshapeposed_{J_{3D}}(\imageset,\set{J}, i) = \Jopenpose(N_{3D}(\imageset,\set{J}, i))
\end{equation}
$\Jopenpose$ has been trained to output $25$ joint locations consistent with the \textsc{Body\_25}~\cite{openpose} keypoint ordering. 
The estimated posed mesh $N_{3D}$ can be rendered in uniform color with the image formation function $R(\cdot)$ paramerized by camera $c$:
\begin{equation}
    \label{eqn:img-formation}
    \outshapeposed_{2D}(\imageset,\set{J}, i) = R_c(\outshapeposed_{3D}(\imageset,\set{J}, i))
\end{equation}
Similarly, we can project the the joints $N_{J_{3D}}$ to the image plane by perspective projection $\pi$:
\begin{equation}
    \outshapeposed_{J_{2D}}(\imageset,\set{J}, i) = \pi_c(\outshapeposed_{J3D}(\imageset,\set{J}, i))
\end{equation}
All these operations are differentiable, which we can conveniently use to formulate suitable loss functions.

\subsection{Loss functions}
\label{sec:losses}
Our architecture permits two sources of supervision: (i) 3D supervision (in our experiments, from synthetic data derived by fitting SMPL+D to static scans), and (ii) 2D supervision from video frames alone.
In this section, we discuss different loss functions used to train the predictors $\netstar$.

\textbf{Losses on body shape and pose}
For a paired sample in the dataset $\{(\imageset, \set{J}), (\shape, \offsets, \poseset, \transset) \}$ we use the following losses between our estimated model $\outshapeposed_{3D}$ and the ground truth model $\smpl(\cdot)$ scan:
\vspace{-1mm}
\begin{itemize}
    \item Per-vertex loss in the canonical T-pose $\vec{0}_{\pose}$. This loss provides a useful 3D supervision on shape independently of pose:
\end{itemize}
\begin{equation}
    \loss_\outshapetpose = || \outshapetpose(\imageset) - \smpl(\shape,\vec{0}_{\pose},\offsets) ||^2
\end{equation}
\begin{itemize}
 \item Per-vertex loss in posed space. This loss supervises both pose and shape on the Euclidean space:
\end{itemize}
\begin{equation}
	\small
    \loss_{\outshapeposed_{3D}} = \sum_{i=0}^{F-1} || \outshapeposed_{3D}(\imageset,\set{J}, i) - \smpl(\shape,\pose_i,\offsets) ||^2
\end{equation}
\begin{itemize}
\item Silhouette overlap:
\end{itemize}
\begin{equation}
    \label{eq:imageProjectionLoss}
    \small
    \loss_{\outshapeposed_{2D}} = \sum_{i=0}^{F-1} || R_c(\outshapeposed_{3D}(\imageset, \set{J},i)) - b(\image_i) ||^2,
\end{equation}
where $b(\image_i)$ is the binary segmentation mask and $R_c$ is the image formation function defined in Eq.~\ref{eqn:img-formation}. 
$\loss_{\outshapeposed_{2D}}$ is a weakly supervised loss as it does not require 3D annotations and $b(\image_i)$ can be estimated directly from RGB images. 
In the experiments, we investigate whether such self-supervised loss can reduce the amount 3D supervision required (see~\ref{sec:Type of supervision}). Additionally, we show that $\outshapeposed_{2D}$ can be used at test time to refine the bottom-up predictions and capture instance specific details in a top-down manner (see~\ref{sec:Scene-specific optimization}).
\begin{itemize}
\item Per-vertex SMPL undressed body loss: %
\end{itemize}
The aforementioned losses only penalize the final SMPL+D 3D shape. It is useful to include an "undressed-body" $(\hat{\outshapetpose})$ loss to force  the shape parameters $\shape$ to be close to the ground truth
\begin{equation}
    \loss_{\hat{\outshapetpose}} = || \hat{\outshapetpose}(\imageset) - \smpl(\shape,\vec{0}_{\pose},\vec{0}_{\offsets}) ||^2
\end{equation}
\begin{equation}
    \hat{\outshapetpose}(\imageset) = \template + \offsetfun_s(\net^{\shape}(\imageset)),
\end{equation}
where $\vec{0}_{\offsets}$ are vectors of length $0$. This also prevents that the offsets $\offsets$ explain the overall shape of the person. 

\textbf{Pose specific losses.} In addition to the posed space $\loss_{\outshapeposed_{3D}}$ and silhouette overlap $\loss_{\outshapeposed_{2D}}$ losses, we train for the pose using a direct loss on the predicted parameters $\loss_{\pose,\trans}$ %
\begin{equation}
	\small
    \loss_{\pose,\trans} = \sum_{i=0}^{F-1} \left( || \mat{R}( \net^{\pose_i}) - \mat{R}(\pose_i) ||^2 + || \net^{\trans_i} - \trans_i ||^2 \right),
\end{equation}
where $\mat{R}$ are vectorized rotation matrices of the 24 joints. Similar to \cite{omran2018neural, Lassner, pavlakos2018humanshape}, we use differentiable SVD to force the predicted matrices to lie on the manifold of rotation matrices. This term makes the pose part of the network converge faster.

\textbf{Losses on joints.} We further regularize the pose training by imposing a loss on the joints in Euclidean space:
\begin{equation}
    \small
    \loss_{J_{3D}} = \sum_{i=0}^{F-1} || \outshapeposed_{J_{3D}}(\imageset,\set{J}, i) - \Jopenpose(\smpl(\shape,\pose_i,\offsets)) ||^2
\end{equation}
Similar to the 2D image projection loss on model $\loss_{\outshapeposed_{2D}}$ (Eq. \ref{eq:imageProjectionLoss}), we also have a weakly supervised 2D joint projection loss $\loss_{J_{2D}}$
\begin{equation}
    \footnotesize
    \loss_{J_{2D}} = \sum_{i=0}^{F-1} || \outshapeposed_{J_{2D}}(\imageset,\set{J}, i) - \pi_c(\Jopenpose(\smpl(\shape,\pose_i,\offsets))) ||^2.
\end{equation}
\subsection{Instance-specific top-down optimization} \label{sec:Scene-specific optimization}
The bottom-up predictions of the neural model can be refined top-down at test time to capture instance specific details. It is important to note that this step requires no 3D annotation as the network fine-tunes using only 2D data.
Specifically, at the test time, given a subject's images $\imageset$ and 2D joints $\set{J}$ we optimize a small set of layers in $\netstar$ using image and joint projection losses $\loss_{\outshapeposed_{2D}}, \loss_{J_{2D}}$ (see~\ref{sec:experiments_setup}). By fixing most layers of the network and optimizing only latent layers, we find a compromise between the manifold of shapes learned by the network and new features, that have not been learned. 
We further regularize this step using Laplacian smoothness, face landmarks, and symmetry terms from \cite{alldieck2018video, alldieck2018detailed}. Table~\ref{tab:error_numerical} illustrates the performance of the pipeline before and after optimization (see~\ref{sec:numerical_evaluation}, \ref{sec:parameter_analysis}).

\section{Experiments}
\label{sec:experiments}
The following section focuses on the evaluation of our method.
In Sec.~\ref{sec:experiments_setup} we introduce technical details of the used dataset and network architecture.
The following sections describe experiments for quantitative and qualitative evaluation as well as ablation and parameter analysis.

\begin{figure}
	\centering
	\includegraphics[width=0.9\columnwidth]{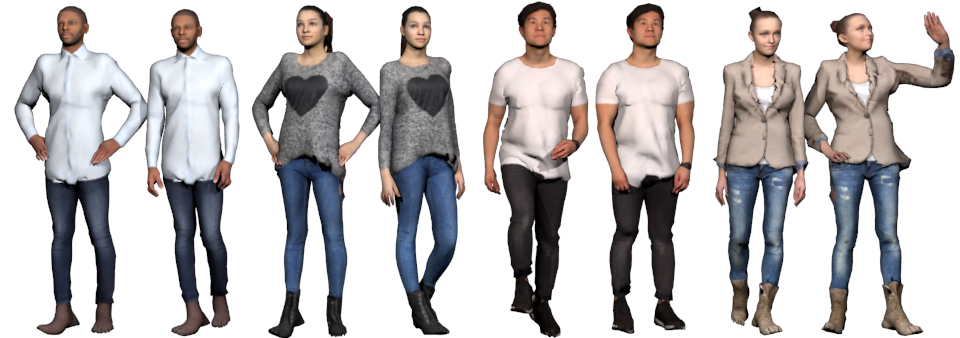}
	\caption{Sample scans from the \emph{LifeScans} dataset.}
	\label{fig:scan_samples}
	\vspace{-4mm}
\end{figure}

\subsection{Experimental setup}
\label{sec:experiments_setup}

\paragraph{Dataset.}
To alleviate the lack of paired data, we use 2043 static 3D scans of people in clothing.
We purchased 163 scans from renderpeople.com and 54 from axyz-design.com. 1826 scans were kindly provided from Twindom (https://web.twindom.com/).
Unfortunately, in the 2043 there is not enough variation in pose and shape to learn a model that generalizes. Hence, we generate synthetic 3D data by non-rigidly registering SMPL+D to each of the scans. This allows us to change the underlying body shape and pose of the scan using SMPL, see Fig.~\ref{fig:scan_samples}. Like~\cite{alldieck2018video}, we focus on a cooperative scenario where the person is turning around in front of the camera. Therefore, we animate the scans with turn-around poses and random shapes and render video sequences from them. We call the resulting dataset~\emph{LifeScans}, which consists of rendered images \emph{paired} with 3D animated scans in various shapes and poses. Since the static scans are from real people, the generated images are close to photo-realistic, see Fig~\ref{fig:scan_samples}. 
To prevent overfitting, we use semantic segmentation together with keypoints as intermediate image representation, which preserve shape and pose signatures while abstracting away appearance. This reduces the amount of appearance variation required for training. 
To be able to render synthetic semantic segmentation, we first render the LifeScans subjects from different viewpoints and segment the output with the method of \cite{gong2018instance}. Then we project the semantic labels back in the SMPL texture space and fuse different views using graph cut-based optimization.
This final step enables full synthetic generation of paired training data.

\vspace{-4.5mm}
\paragraph{Scale ambiguity.} Scale is an inherent ambiguity in monocular imagery.
Three factors determine the size of an object in an image: distance to the camera, camera intrinsics, and the size of the object. 
As it is not possible to decouple this ambiguity in a monocular set-up with moving objects, we fix two factors and regress one.
In other works \cite{omran2018neural,kanazawa2018endtoend,pavlakos2018humanshape} authors have assumed fixed distance to the camera.
We cannot make this assumption, as we leverage multiple images of the same subject, where the distances to the camera may vary.
Consequently, we fix the size of the subject to average body height. Precisely, we make SMPL height independent, by multiplying the model by $1.66$m divided by the y-axis distance of vertices describing ankles and eyes. Finally, we fix the focal length to sensor height.

\vspace{-4.5mm}
\paragraph{Network architecture.}
\begin{figure*}
	\centering
    \begin{overpic}[width=0.95\textwidth]{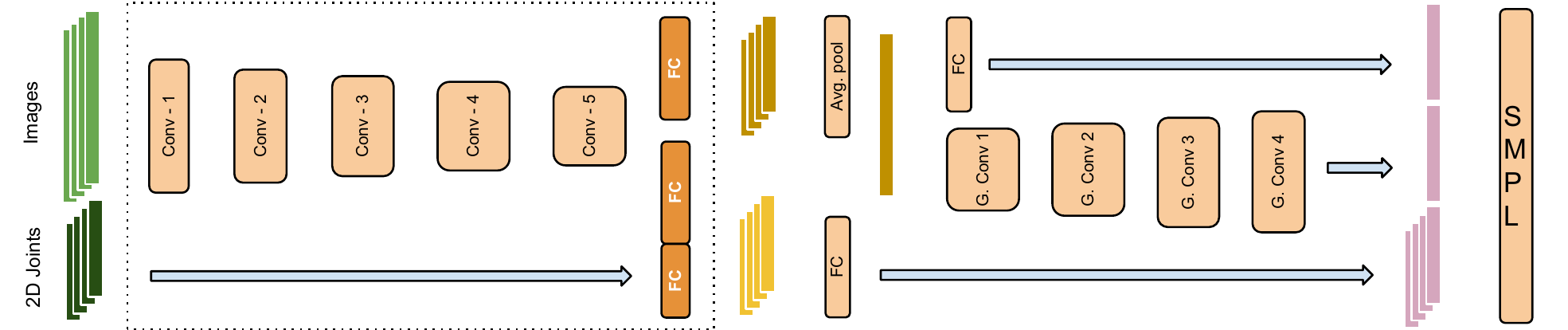}
        \put (92.5,16.5) {\rotatebox[origin=c]{90}{$\shape$}}
        \put (92.5,10.5) {\rotatebox[origin=c]{90} {$\offsets$}}
        \put (92.5,4) {\rotatebox[origin=c]{90} {$\poseset, \transset$}}
        \put (50,5) {\rotatebox[origin=c]{90} {$\latentPose$}}
        \put (57.75,14) {\rotatebox[origin=c]{90} {$\latentCode$}}
    \end{overpic}
	\caption{Network architecture: Our bottom-up inference network first encodes the input (semantically segmented images $\imageset$ and 2D joints $\set{J}_{2D}$) into a decoupled, pose dependent $\latentPose$ and pose invariant $\latentCode$ latent space. The pose branch subsequently infers per-frame pose and translation parameters $\poseset$ and $\transset$ from $\latentPose$. The shape branch infers the body shape $\shape$ and free-form deformations $\offsets$ in T-pose from $\latentCode$. We use a graph convolution based decoder, to learn per-vertex offsets $\offsets$. The entire model is end-to-end trainable. The orange FC layers and the final graph convolution layer can be fine-tuned at test time to better model instance-specific details (see Sec. \ref{sec:Scene-specific optimization}).}
	\label{fig:network}
	\vspace{-2mm}
\end{figure*}
In the following we describe details of the convolutional neural network $\netstar$. An overview is given in Fig.~\ref{fig:network}.
The input to $\netstar$ is a set of $1080$x$1080$px semantically segmented images $\imageset$ and corresponding 2D joint locations $\set{J}$. %
$\netstar$ encodes each image $\image_i$ with a set of five, $3$x$3$ convolutions with ReLU activations followed by $2$x$2$ max-pooling operations into a pose invariant latent code $\latentCode_i$.
In our experiments we fixed the size of $\latentCode_i$ to $20$.
The \emph{pose branch} maps both joint detections $\joints_i$ and output of the last convolutional layer to a vector of size $200$ and finally to the pose-dependent latent code $\latentPose_i$ of size $100$ via fully connected layers.
The \emph{shape branch} aggregates pose invariant information across images and computes mean $\latentCode$. Note that this formulation allows us to aggregate pose-dependent and invariant information across an arbitrary and varying number of views. The shape branch goes on to predict SMPL shape parameters $\shape$ and free-form deformations $\offsets$ on the SMPL mesh. $\shape$ is directly calculated from $\latentCode$ with a linear layer.
In order to predict per-vertex offsets from the latent code $\latentCode$, we use a four-step graph convolutional network with Chebyshev filters and mesh upsampling layers similar to \cite{COMA:ECCV18}. Each convolution is followed by ReLU activation. We prefer a graph convolutional network over a fully connected decoder due to memory constraints and in order to get structured predictions.

\vspace{-4.5mm}
\paragraph{Training scheme.}
The proposed method, including rendering, is fully differentiable and end-to-end trainable. Empirically we found it better to train the pose branch before training the shape branch. Thereafter, we optimize the network end-to-end.
We use a similar training schedule for our pose branch as \cite{pavlakos2018humanshape}, where we first train the network using losses on the joints and pose parameters ($\loss_{J_{3D}}, \loss_{\pose,\trans}$) followed by training using losses on the vertices and pose parameters ($\loss_{\outshapeposed_{3D}}, \loss_{\pose,\trans}$). 
We also experiment with various training schemes, and show that weakly supervised training can significantly reduce the dependence on 3D annotated data (see Sec.~\ref{sec:Type of supervision}). For that experiment, we train the model with alternating full ($\loss_\outshapetpose, \loss_{\hat{ \outshapetpose}},  \loss_{\outshapeposed_{3D}}, \loss_{J_{3D}}$) and weak supervision ($\loss_{\outshapeposed_{2D}}, \loss_{J_{2D}}$). 
During instance-specific optimization we keep most layers fixed and only optimize latent pose $\latentPose$, latent shape $\latentCode$ and the last graph convolutional layer, that outputs free-form displacements $\offsets$.

\subsection{Numerical evaluation}
\label{sec:numerical_evaluation}
\begin{figure*}
	\includegraphics[width=\textwidth]{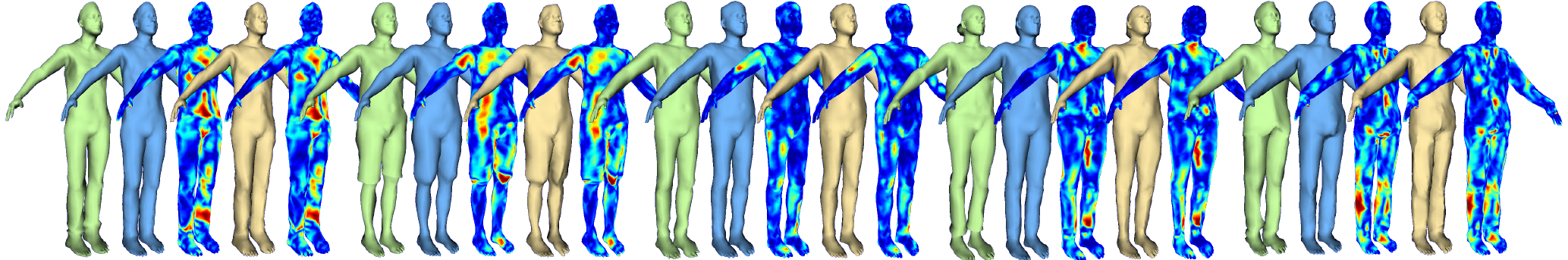}
	\caption{Results from LifeScans in comparison to ground truth shapes (green). We show results computed with ground truth poses (blue) and results of the full method (yellow) with corresponding error heatmaps with respect to ground truth shapes (red means $\geq2$cm).}
	\label{fig:heatmaps}
	\vspace{-2mm}
\end{figure*}
\begin{figure*}
	\scriptsize
	\centering
	\begin{overpic}[width=0.95\textwidth]{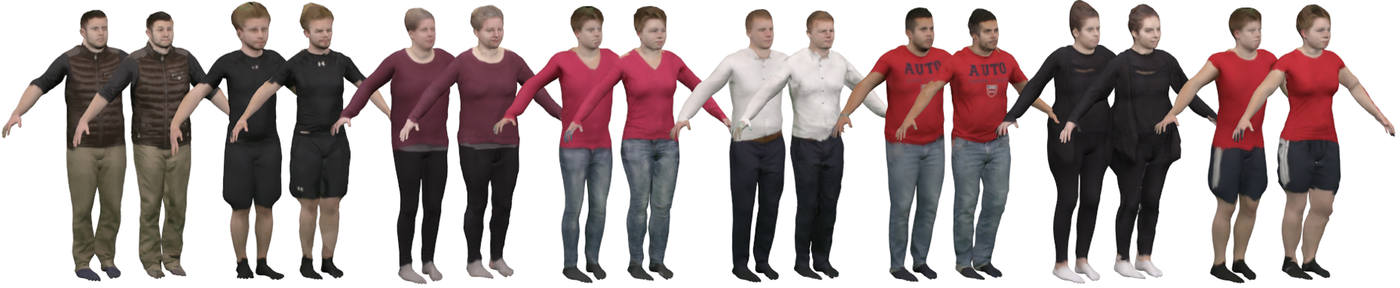}
		\put (3.8,1.5) {a)}
		\put (8.6,1.5) {b)}
	\end{overpic}
	\caption{Comparison to the state-of-the-art optimization based method \cite{alldieck2018video}. Their method (a) uses 120 frames, while ours (b) only uses 8 images and is several magnitudes faster.}
	\label{fig:peoplesnapshot}
			\vspace{-2mm}
\end{figure*}

We quantitatively evaluate our method on a separated test set of the LifeScans dataset containing 55 subjects. We use $F=8$ semantic segmentation images and 2D poses as input and optimize the results for a maximum budget of $10$ seconds.
All results have been computed without intensive hyper-parameter tuning.
To quantify shape reconstruction accuracy, we adjust the pose of the estimation to match the ground truth, following~\cite{zahng2017shapeundercloth, Bogo:ICCV:2015}. 
This disentangles errors in pose from errors in shape and allows to quantify shape accuracy.
Finally, we compute the bi-directional vertex to surface distance between scans and reconstructions. We report mean errors in millimeters (mm) across the test set in Tab.~\ref{tab:error_numerical}.
We differentiate between \emph{full method} and \emph{ground truth (GT) poses}.
Full method refers to our method as described in Sec.~\ref{sec:experiments_setup}.
The latter is a variant of our method that uses ground truth poses, which allows to study the effect of pose errors.
In Fig.~\ref{fig:heatmaps} we display subjects in the test set for both variants along with per-vertex error heatmaps.
Visually the results look almost indistinguishable, which is corroborated by the fact that the numerical error increases only by $\approx 1$mm between GT and predicted pose models. This demonstrates the robustness of our approach.
We show more examples with the corresponding texture for qualitative assessment in Fig.~\ref{fig:teaser}. The textures have been computed using graph cut-based optimization using semantic labels as described in \cite{alldieck2018detailed}.

\begin{table}
	\centering
	{\footnotesize
		
\begin{tabular}{l||c|c}
	& Before optimization & After optimization \\  \hline
	Full Pipeline & 4.47 $\pm$4.45 & 4.00 $\pm$3.94 \\
	GT Poses & 4.47 $\pm$4.41 & 3.17 $\pm$3.41 \\
	\hline\end{tabular}
		\vspace{2mm}
	}
	\caption{Mean vertex error (mm) of 55 test samples computed on $F=8$ input images. The \emph{full method} with inferred poses produces comparable results to using \emph{GT poses}. Both variants gain accuracy from subsequent optimization.}
	\label{tab:error_numerical}
	\vspace{-2.5mm}
\end{table}

\subsection{Analysis of key parameters}
\label{sec:parameter_analysis}
Our method comes with two key hyper-parameters, namely number of input images $F$, and number of optimization steps.
In the following section, we study these parameters and how they affect the performance of our approach. We also justify our design choices.

\begin{figure}
		\vspace{-2mm}
	\centering
	\includegraphics[width=\columnwidth]{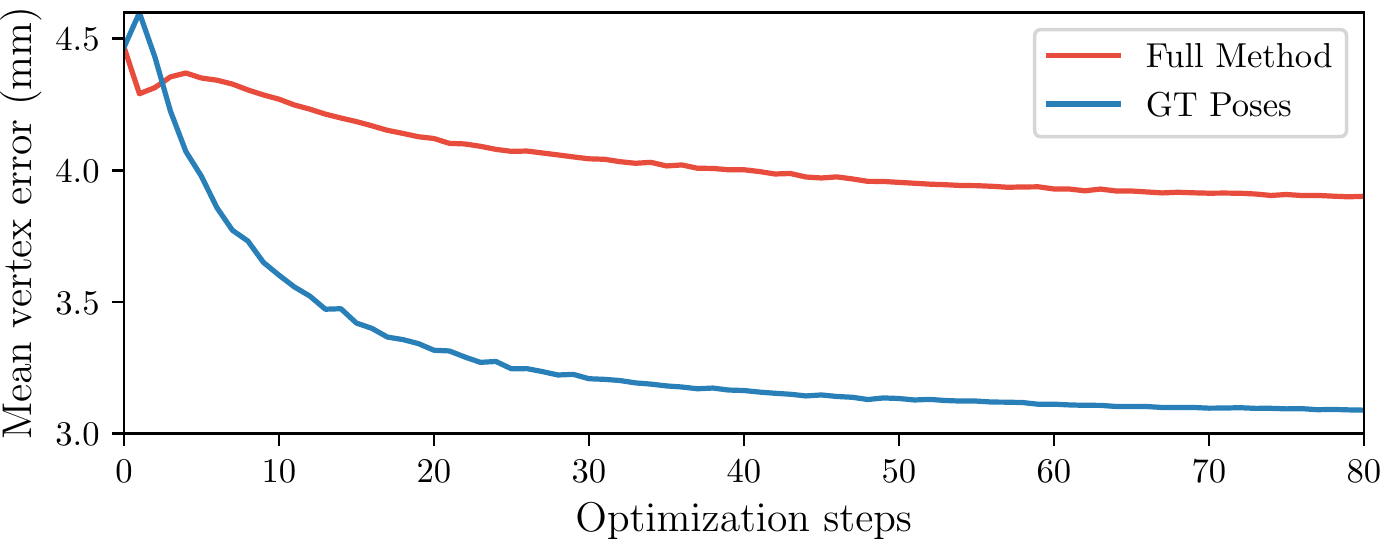}
	\caption{Error decrease of the test set with increased number of optimization steps computed on $F=8$ input images.} %
	\label{fig:error_vs_opt}
	
	\vspace{1mm}
	\includegraphics[width=\columnwidth]{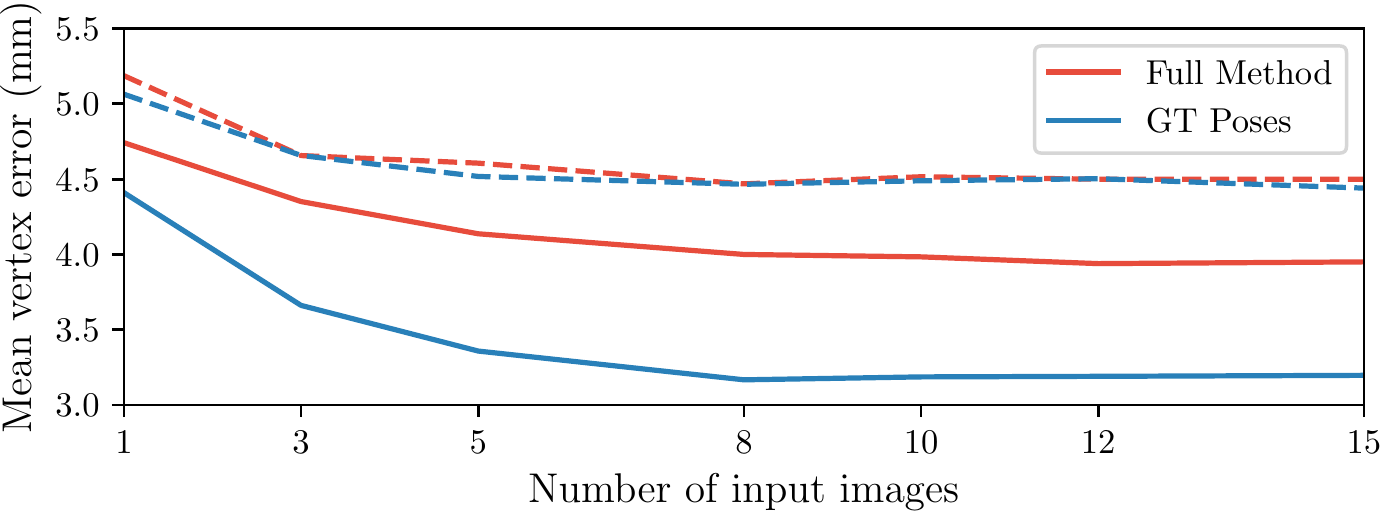}
	\caption{Error development on the test set with increased number of input views $F$ before (dashed) and after optimization (solid). Optimization has been limited by a time budget of $10$s allowing very few gradient steps for large numbers of views, which explains why the error plateaus for more than $8$ views.}
	\label{fig:error_vs_num}
	\vspace{-2mm}
\end{figure}

Fig.~\ref{fig:error_vs_opt} illustrates the performance of our method with growing number of optimization steps.
While the performance gain saturates at around $70-80$ steps, we use $25$ steps in following experiments as a compromise between accuracy and speed. For the case of $F=8$ input images optimization for $25$ steps takes $\approx10$s on a single Volta V100 GPU. We believe $10$s is a practical waiting time and a good compromise for many applications. Therefore we fix the time budget to $10$s for the following experiments.

Including more input views at test time can potentially improve the performance of the method.
However, in practice, this means more data pre-processing and longer inference times.
Fig.~\ref{fig:error_vs_num} illustrates the performance with different number of input images.
Perhaps surprisingly, the performance saturates already at around 5 images before optimization.
After optimization, the error saturates at around 8 images.
While more images potentially means  better supervision, we cannot see improved results for optimization on many images.
This can be explained with the fixed time budget in this experiment, where more images mean fewer optimization steps.
While we could potentially use fewer images, we found $F=8$ views as a practical number of input views.
This has the following reason: A calculated avatar should not only be numerically accurate but also visually appealing.
Results based on more number of views show more fine details and most importantly allow accurate texture calculation. %

\subsection{Type of supervision} \label{sec:Type of supervision}
Since videos are easier to obtain than 3D annotations, we evaluate to which extent they can substitute full 3D supervision to train our network.
To this end, we split the LifeScans dataset. One part is used for full supervision, the other part is used for weak supervision in form of image masks and 2D keypoints.
All forms of supervision can be synthetically generated from the LifeScans dataset.
We train $\netstar$ with 10\%, 20\%, 50\%, and 100\% full supervision and compare the performance on the test set in Tab.~\ref{tab:type_of_supervision}.
In order to factor out the effect of problematic poses during the training, we used ground truth poses in this experiment.
The results suggest that $\netstar$ can be trained with only minimal amount of full supervision, given strong pose predictions.
The performance of the network decreases only slightly for less than 100\% full supervision.
Most interestingly, the results are almost identical for 10\%, 20\%, and 50\% full supervision. 
This experiment suggests that we could potentially improve performance by supervising our model with additionally recorded videos. We leave this for future work.

\begin{table}
\centering
{\footnotesize

\begin{tabular}{r||c|c}
	& Before optimization & After optimization \\ \hline
	100\% & 4.47 $\pm$4.41 & 3.17 $\pm$3.41 \\
	50\% & 4.57 $\pm$4.52 & 3.19 $\pm$3.43 \\
	20\% & 4.74 $\pm$4.65 & 3.29 $\pm$3.53 \\
	10\% & 4.73 $\pm$4.56 & 3.46 $\pm$3.62 \\
	\hline\end{tabular}
\vspace{2mm}
}
\caption{Mean vertex error (mm) of 55 test samples with different amount of \emph{full supervision} during training of the shape branch. $\netstar$ can be trained with only 10\% full supervision with minimal accuracy lose.}
\label{tab:type_of_supervision}
\vspace{-3mm}
\end{table}

\subsection{Qualitative results and comparisons}

We qualitatively compare our method against the most relevant work~\cite{alldieck2018video} on their \emph{PeopleSnapshot} dataset. While their method leverages 120 frames, we still use $F=8$ frames for our reconstructions. 
For a fairer comparison, we optimize for $\approx 20$s in this experiment. This is still several magnitudes faster than the $122$min needed by \cite{alldieck2018video}. Their method needs $2$ minutes for shape optimization plus $1$ minute per frame for the pose.
In Fig.~\ref{fig:peoplesnapshot} we show side-by-side comparison to~\cite{alldieck2018video}.
Our results are visually still on par while requiring a fraction of the data.

\begin{figure}
	\vspace{-4mm}
	\scriptsize
	\centering
	\begin{overpic}[width=0.95\columnwidth]{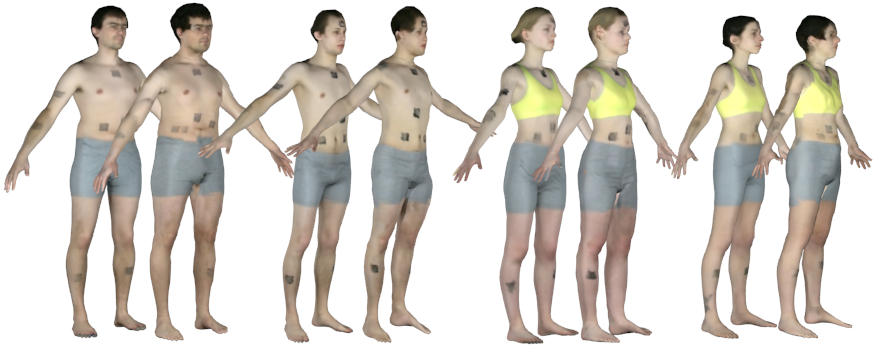}
		\put (5,3.2) {a)}
		\put (15.3,3.2) {b)}
	\end{overpic}
	\caption{Comparison to the RGB-D method \cite{Bogo:ICCV:2015} (a). Our method (b) is visually on par, despite using only 8 RGB images as input.}
	\label{fig:kinectcap}
	\vspace{-3.5mm}
\end{figure}

We also compare our method against~\cite{Bogo:ICCV:2015}, a RGB-D based optimization method.
Their dataset displays subjects in minimal clothing rotating in front of the camera in T-pose.
Unfortunately, the semantic segmentation network is not able to successfully segment subjects in minimal clothing.
Therefore we sightly change the set-up for this experiment. We segment their dataset using the semi-automatically approach~\cite{caelles2017oneshot} and re-train our predictor to be able to process binary segmentation masks.
Additionally, we augment the LifeScans dataset with T-poses.
We show side-by-side comparisons in Fig.~\ref{fig:kinectcap}.
Again our results are visually similar, despite the use of less and only monocular data.

\section{Discussion and Conclusion}
\label{sec:conclusion}
We have proposed a novel method for automatic 3D body shape estimation from only $1-8$ frames of a monocular video of a person moving.
Our \emph{Octopus} model predicts mesh-based pose invariant shape and per-image 3D pose from a flexible number of views.
Experiments demonstrate that the feed-forward predictions are already quite accurate ($4.5$mm), but often lack detail and do not perfectly overlap with the input images. 
This motivates refining the estimates with top-down optimization against the input images. Refining brings the error down to $4$mm and aligns the model with the input image silhouettes, which allows texture mapping. 
In summary, we improve over the state-of-the-art in the following aspects:
Our method allows, for the first time, to estimate full body reconstructions of people in clothing in a fully automatic manner. We significantly reduce the number of needed images at test time, and compute the final result several magnitudes faster than state-of-the-art (from hours to seconds). 
Extensive experiments on the LifeScans dataset demonstrate the performance and influence of key parameters of the predictor.
While our model is independent on the number of input images and can be refined for different numbers of optimization steps, we have shown that using $8$ views and refining for $10$ seconds are good compromises between accuracy and practicability. %
Qualitative results on two real-world datasets demonstrate generalization to real data, despite training from synthetic data alone.

Future work should enable the proposed method for scenarios where the subject is not cooperating, for example from Youtube videos, or legacy movie material. 
Furthermore, clothing with geometry far from the body, such as skirts and coats or hairstyles like ponytails will require a different formulation.

By enabling fully automatic 3D body shape reconstruction from a few images in only a few seconds, we prepare the ground for wide-spread acquisition of personalized 3D avatars.
People are now able to quickly digitize themselves using only a webcam and can use their model for various VR and AR applications.

{\small
	\vspace{1mm}
	\noindent
	\textbf{Acknowledgments}\\
	The authors gratefully acknowledge funding by Deutsche Forschungsgemeinschaft (DFG German Research Foundation) from projects MA2555/12-1 and 409792180. We would like to thank Twindom for providing us with the scan data. Another thanks goes to Verica Lazova for great help in data processing.
}

{\small
\bibliographystyle{ieee}
\bibliography{egbib}
}

\end{document}